\documentclass[conference]{IEEEtran}
\ifCLASSINFOpdf
\else
\fi
\usepackage{graphicx}
\usepackage{amsmath}
\graphicspath{{C:/images/}}
\begin{document}

%
\title{One-Shot Concept Learning by Simulating Evolutionary Instinct Development}

\author{\IEEEauthorblockN{Abrar Ahmed}
\IEEEauthorblockA{South Forsyth High School\\
Cumming, Georgia 30041\\
}
\and
\IEEEauthorblockN{Anish Bikmal}
\IEEEauthorblockA{South Forsyth High School\\
Cumming, Georgia 30041\\
}
}


%


\maketitle

\begin{abstract}
Object recognition has become a crucial part of machine learning and computer vision recently. The current approach to object recognition involves Deep Learning and uses Convolutional Neural Networks to learn the pixel patterns of the objects implicitly through backpropagation. However, CNNs require thousands of examples in order to generalize successfully and often require heavy computing resources for training. This is considered rather sluggish when compared to humans’ ability to generalize and learn new categories given just a single example. Additionally, CNNs make it difficult to explicitly programmatically modify or intuitively interpret their learned representations. 
We propose a computational model that can successfully learn an object category from as few as one example and allows its learning style to be tailored explicitly to a scenario. Our model decomposes each image into two attributes: shape and color distribution. We then use a Bayesian criterion to probabilistically determine the likelihood of each category. The model takes each factor into account based on importance and calculates the conditional probability of the object belonging to each learned category. Our model is not only applicable to visual scenarios, but can also be implemented in a broader and more practical scope of situations such as Natural Language Processing as well as other places where it is possible to retrieve and construct individual attributes. Because the only condition our model presents is the ability to retrieve and construct individual attributes such as shape and color,  it can be applied to essentially any class of visual objects no matter how large or small.

\end{abstract}


%
\IEEEpeerreviewmaketitle

\section{Introduction}
Recognition is one of the most remarkable features of biological cognition. This important function is not only present in humans but also in animals with less developed cognitive abilities such as birds. The ability to recognize surroundings, prey, and potential mates is critical to the survival of any creature regardless of environment or function. The recognition capabilities that these animals possess are unique in the way that they require very few training examples in order to learn a new category or concept. For example, after a chick sees its guardian capture a fish and eat it, it will learn that a fish means food. Furthermore, using this learned concept, the bird may associate similar organisms like tadpoles to also be food. People and animals can use concepts in richer ways than conventional computational models or algorithms - for action, imagination, an explanation. 
Additionally, CNNs make it difficult to explicitly programmatically modify or intuitively interpret their learned representations. 
Humans and animals require few examples to learn when compared to even the latest machine models involving deep learning. Even though deep learning models are based on the biological brain, they require many more examples because they lack an important feature that animals are born with: instinct. Animals come pre-programmed with evolutionary instinct that allows them to learn concepts much faster while Deep Learning models often initialize their weights randomly. This gives animals a huge advantage when it comes to learning concepts and representations quickly [10]. Natural selection has gifted them with basic concepts that their ancestors learned through trial and error the hard way.
	We present a computational model that can successfully learn an object category from as few as one example and allows its learning style to be tailored explicitly to a scenario. Our model compensates for the evolutionary instincts that animals are born with. Just as different animals have different instincts, the instincts of our model are adaptable to different scenarios and tasks. Our model “is born with” an instinctive ability to differentiate objects based on color and shape. The model takes each factor into account based on importance and calculates the conditional probability of the object belonging to each learned category through a Bayesian logistic regression.

\includegraphics[scale=.26]{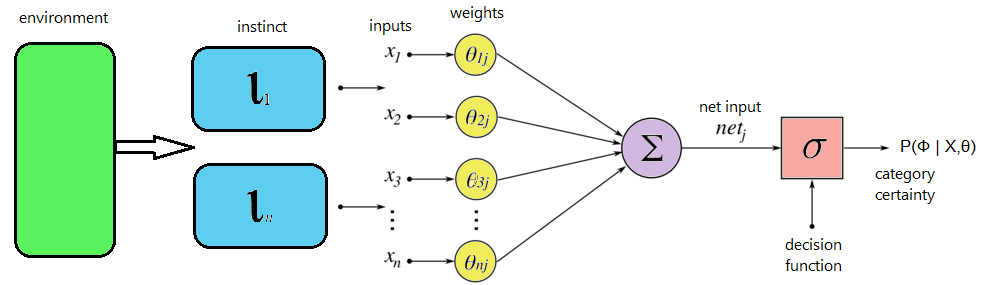}

\section{One-Shot Learning}
One-Shot Learning can be defined as the ability to use concepts in richer ways than conventional computational models or algorithms - for action, imagination, an explanation. It is a new relatively new topic in the field of machine learning and computer vision. The goal of our model is to outperform traditional deep learning models by implementing this concept of one-shot learning in the field of object recognition. Our model successfully recognized a set of fruits of vegetables with "one-shot" and it can use just one image to recognize not only fruits, but any image with the parameters of color and shape - a huge class of objects. 

\section{Techniques}

\subsection{Dataset}
30 images of one category and 3 images of all other categories were obtained for training and testing the classifier.

\subsection{Pre - processing}	
Each image was scaled to the same size. Canny Edge Detection was performed on each image to obtain an edge outline of each image [2]. Edge outlines were then fed into a binary thresholding function to ensure pixel values were either 0 or 255. The locations of pixels with a value of 255 were stored in an n by 2 pixel coordinate array. The coordinates are then centered by subtracting the respective axis median from each column of the array. The height of the outline of each image is then scaled up to the height of the image with the greatest height to prevent errors due to the magnification of the image by multiplying the corresponding pixel coordinate array by a scale factor. The image outlines are used as masks on the original image to locate a color sample on the surface of each object. The color sample is processed to remove the alpha transparency values. The color sample in RGB color sample is then projected into the YUV color space because it better represents human perceived color similarity as demonstrated by Podpora et al. [8]. The three dimensional color sample matrix is then averaged to produce a three dimensional average color vector for each image.  
\vspace*{0cm}
\includegraphics[scale=.7]{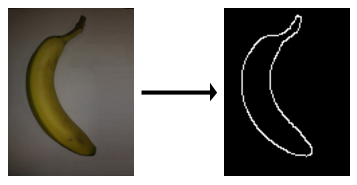}
\\
(Figure 1) Using Canny edge detection (Canny et al.) [2] to obtain outline of image.
\\
\\
$\begin{bmatrix}$
${.299} & {.587} & {.114} \\ 
{-.14713} & {-.28886} & {.436} \\
{.615} & {-.51499} & {-.10001}$
$\end{bmatrix}$
$\times$
$\begin{bmatrix}$
$
{R}\\ 
{G}\\
{B}$
$\end{bmatrix}$
= 
$\begin{bmatrix}$
$
{Y'}\\ 
{U}\\
{V}$
$\end{bmatrix}$ \\
\\

$\longrightarrow$
$\sqrt{(Y'_{2}-Y'_{1})^{2} + (U_{2}-U_{1})^{2} + (V_{2}-V_{1})^{2}}$
= $\epsilon_{i}$
\\
\\
Converting RGB values to YUV using a coefficient matrix to calculate Euclidean distance.

\subsection{Procedure}
Modified Hausdorff Distance developed by Dubuisson et al. [1] is then used as a morphological similarity metric to gauge the difference between two image outlines. To determine chromatic similarity, the Euclidean distance is calculated between the YUV average color vectors giving Delta E, the color difference. Feature scaling is then used on the MHD and Delta E values based on their maximum ranges to yield similar ranges for both. A Bayesian Logistic Regression model is then trained on the resulting MHD and Delta E values of the 30 bananas and remaining non-banana objects in order to determine the coefficient matrix theta that minimizes the error. The resulting logistic function is then used to output the conditional probability of an object being from a certain category given the MHD and Delta E values of that object and that category. The model then predicts the most probable category of the object. The objects were divided into two sets: objects with distinct color and shape attributes, and objects with nearly identical color and shape features. Because we had 3 images of each object, we were able to make 3P2 = 6 comparisons for each image to image pair. For the first set, $6\times9$ = 54 comparisons were made. For the second set, $6\times8$ = 48 comparisons were made. Three different Deep Neural Networks were trained on the same set using Amazon AWS cloud services and their accuracies were analyzed in comparison to our one shot approach. Additionally, a group of 6 participants were each shown a portion of the images and asked to classify them given the possible categories. The image sets were constructed so that there were duplicates of some objects and absence of others so that participants could not use process of elimination.

\includegraphics[scale=.5]{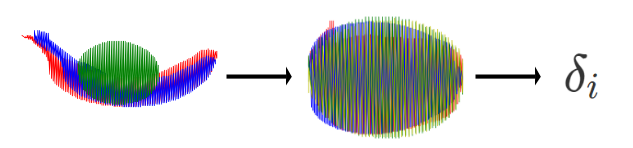}
(Figure 2) Scaling the images to the same width, then using modified Hausdorff distance to find the $\delta_{1}$ value or the "error in shape."
\vspace*{0cm}
\\
$P(\Phi_{i}|\delta_{i},\epsilon_{i})=h_{\theta}(x^{(i)})= \dfrac{1}{1+e^{-\theta^{T} [\delta_{i},\epsilon_{i}]}}$  (Equation 1) \\
\\
$\theta = min_{\theta}J(\theta)$  (Equation 2) \\
\\
$J(\theta) = -\dfrac{1}{m}[\sum\limits_{i=1}^m y^{(i)} log h_{\theta}(x^{(i)})+(1-y^{(i)})log(1-h_{\theta}(x^{(i)}))]$ (Equation 3)
\vspace*{0mm}
\\
Equation 1 uses the intermediate step of the logistic regression to give us the exact probability of a scanned object being an object given the shape (MHD - $\delta_{i}$) and color (Euclidean distance - $\epsilon_{i}$).\\
Equation 2 minimizes the error of the cost function.\\
Equation 3 is the cost function.

\section{Results}
Our model proved to be successful in generalizing concepts using one example, and outperformed conventional methods using deep learning neural networks, and in some cases, it even outperformed humans. We found that the accuracy varied significantly based on the objects that we were using. In fact, we found that the results were strikingly similar to a human’s performance. For example, the model mixed up zucchinis and cucumbers, and bananas and plantains relatively more than it mixed up other objects, yet still performed better than humans at differentiating these nearly identical objects . However, when the “distinct objects” (the objects that are not easily mistaken for others) were tested alone, the accuracy was almost perfect - 98.15 percent. When the “similar objects” (the objects easily mixed up) were tested separately, the accuracy of our model was about 72.92 percent. At first glance this statistic may seem low, but it was more accurate than a human at detecting a “similar fruit.” An average human’s accuracy rate turned out be 100 percent on the distinct objects, outperforming our model by 1.85 percent. However,  the human prediction success rate is significantly less than our model’s accuracy on the identical objects coming in at 62.51 percent. Many subjects frequently mixed up zucchinis and cucumbers, and bananas and plantains just like our model did. 
\includegraphics[scale=.45, angle=270]{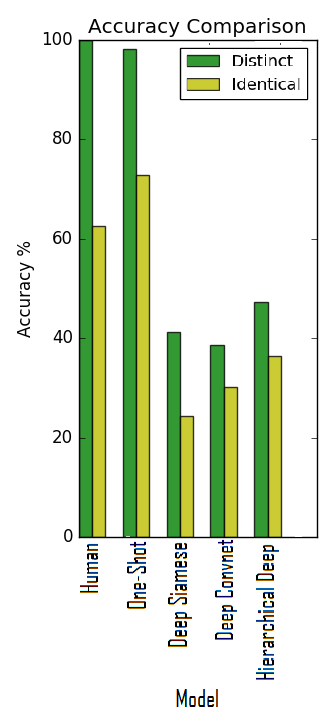}
(Figure 3) Comparison of accuracy rates between humans one-shot(our model), and the three most commonly used deep learning models.\\
\\
Figure 4 is a histogram of the sum of the color and shape errors (after being appropriately weighted) of 30 bananas and 30 non-bananas. Our initial thought was to use some type of standard distribution to fit the data. However, standard distributions like chi-square distributions do not have a distinct gap in the middle of the data. Although this gap prevented us from using a chi-square distribution, it showed that there is a clear distinction between the correct and incorrect objects. We finally decided to use a Bayesian logistic regression fit which is appropriate in our task of classification. It proved to be very accurate, outperforming conventional deep learning models and even outperforming humans on certain tasks. The three deep learning models - Deep Siamese convent, deep convent, and Hierarchical Deep had accuracy rates of approximately 32.37 percent, 34.76 percent, and 40.07 percent respectively when given the same data as our model (can be seen in Figure 3).
Our model demonstrates that object categories can be generalized successfully from few examples using a Bayesian classifier, coming close to the abilities of humans by compensating for the evolutionary instinct that animals are born with. \\
\\
\includegraphics[scale=.6]{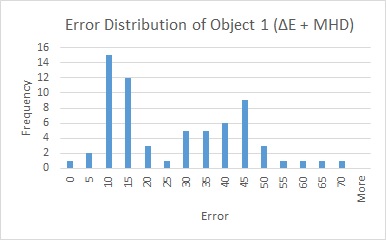}\\
(Figure 4) Histogram of sum of $\delta_{i}$ and $\epsilon_{i}$ to see if a standard distribution will fit the data.
\vspace*{0mm}
\\
\includegraphics[scale=.4]{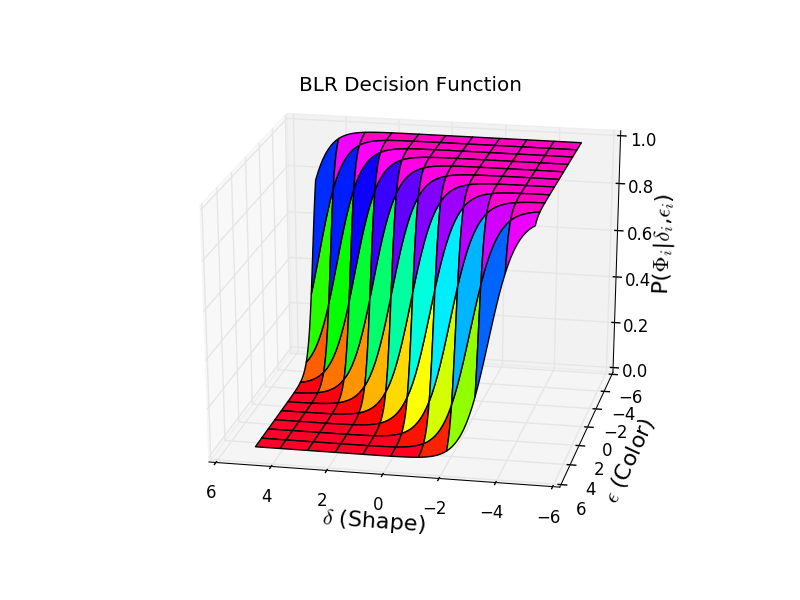}\\
(Figure 5) Bayesian Logistic Regression showing probabilities based on $\delta_{i}$ and $\epsilon_{i}$.
\vspace*{0mm}
\\
\includegraphics[scale=.45]{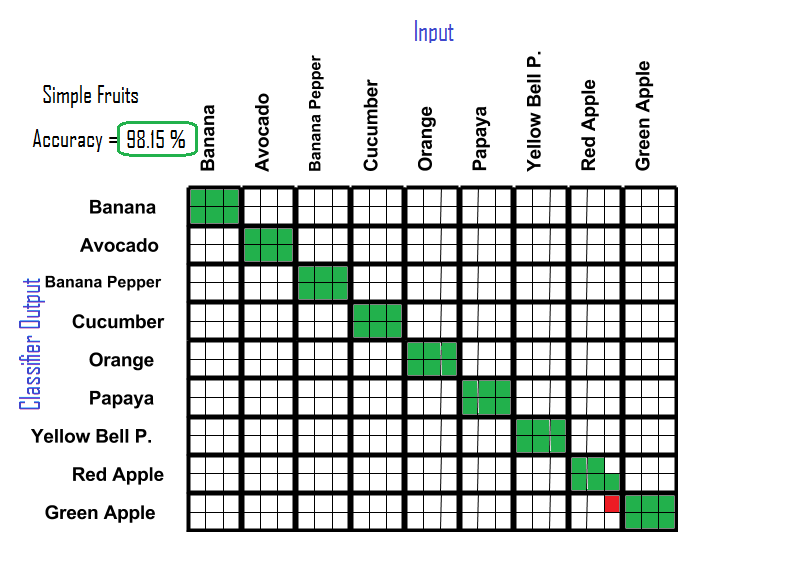}

\includegraphics[scale=.45]{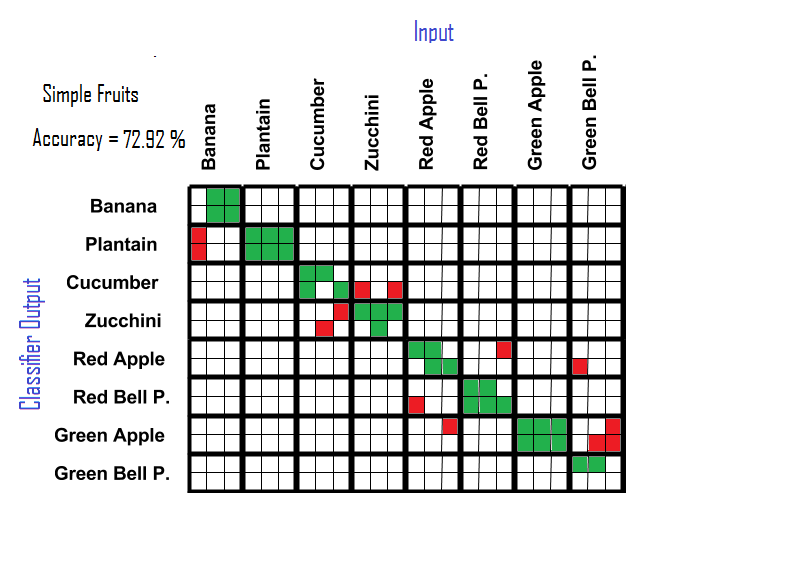}
(Figures 6 and 7) Specific classifier outputs split into two groups (distinct and similar fruits) to analyze data better.

\section{Discussion}
The only condition our model presents to work effectively is the ability to retrieve and construct individual attributes such as shape and color. Therefore it can be applied to a vast variety of classes of visual objects no matter how large or small. Using the individual attributes of shape and color, our model can be applied to help in the recognition of cells, pathogens, DNA etc. Astrocytes, which are star shaped glial cells that act as “the immune system” of the nervous system, perform many functions including biochemical support of endothelial cells that form the blood–brain barrier, provision of nutrients to the nervous tissue, maintenance of extracellular ion balance, and a role in the repair and scarring process of the brain and spinal cord following traumatic injuries [11]. When stained with antibodies to GFAP and vimentin, they appear yellow. From a picture of a tissue from a microscope, our model should be able to successfully identify these astrocytes based on their shape and color. Moreover, number of neurodegenerative diseases like Alexander’s disease are characterized by intra-astrocytic protein aggregates, consisting of mutant GFAP, heat shock protein 27, and αβ-crystallin which can result in a conformational change in the shape of the astrocyte and a change in the GFAP concentration that will result in a color change [12]. Our model will not only recognize which cells are astrocytes, but based on the shape and color changes, it can also recognize which of these astrocytes are diseased. 
An important aspect of our model is that the input parameters, which in our application were color and shape, are fully interchangeable and are able to be chosen differently depending on the scenario. This gives the model a special degree of flexibility and allows it to be expanded to use other individual attributes such as parts, subparts, and spatial relations in classes do not have color as a variable. This will present a method of recognition in classes like characters (alphabets and numerals), and other colorless classes. Our model which takes a one-shot approach, can revolutionize object recognition and can be implemented to essentially any class of visual objects.


\section*{Acknowledgment}

We would like to thank Carol Sikes - high school AP Calculus and Statistics teacher - for assisting us.



%

\end{document}